\pdfoutput=1

\documentclass[11pt]{article}

\usepackage[final]{coling}

\usepackage{times}
\usepackage{latexsym}
\usepackage{amsmath}

\usepackage{float}
\usepackage{placeins}

\usepackage[T1]{fontenc}

\usepackage[utf8]{inputenc}

\usepackage{microtype}

\usepackage{inconsolata}

\usepackage{graphicx}

\title{Why Does ChatGPT ``Delve'' So Much? Exploring the Sources of Lexical Overrepresentation in Large Language Models}

\author{Tom S.\ Juzek \and Zina B.\ Ward\thanks{Conceptually, both authors contributed equally to this work. Tom wrote the code to the paper, which can be accessed at \href{https://github.com/tjuzek/delve}{github.com/tjuzek/delve}.}\\
 Florida State University \\
 \texttt{tjuzek@fsu.edu, zward@fsu.edu}\\}

\begin{document}
\maketitle
\begin{abstract}
Scientific English is currently undergoing rapid change, with words like ``delve,'' ``intricate,'' and ``underscore'' appearing far more frequently than just a few years ago. It is widely assumed that scientists' use of large language models (LLMs) is responsible for such trends. We develop a formal, transferable method to characterize these linguistic changes. Application of our method yields 21 focal words whose increased occurrence in scientific abstracts is likely the result of LLM usage. We then pose ``the puzzle of lexical overrepresentation'': \textit{why} are such words overused by LLMs? We fail to find evidence that lexical overrepresentation is caused by model architecture, algorithm choices, or training data. To assess whether reinforcement learning from human feedback (RLHF) contributes to the overuse of focal words, we undertake comparative model testing and conduct an exploratory online study. While the model testing is consistent with RLHF playing a role, our experimental results suggest that participants may be reacting differently to ``delve'' than to other focal words. With LLMs quickly becoming a driver of global language change, investigating these potential sources of lexical overrepresentation is important. We note that while insights into the workings of LLMs are within reach, a lack of transparency surrounding model development remains an obstacle to such research. 
\end{abstract}

\section{Introduction}
\label{sec:introduction}

Like all human language, Scientific English has changed substantially over time \cite{degaetano2018using, degaetano2018information, bizzoni2020linguistic,menzel2022medical}. New discoveries have fueled (and perhaps been fueled by) the introduction of new lexical items into scientific discourse \cite{degaetano2018using}. 

\begin{figure}[ht]
    \centering
    \includegraphics[width=1\columnwidth]{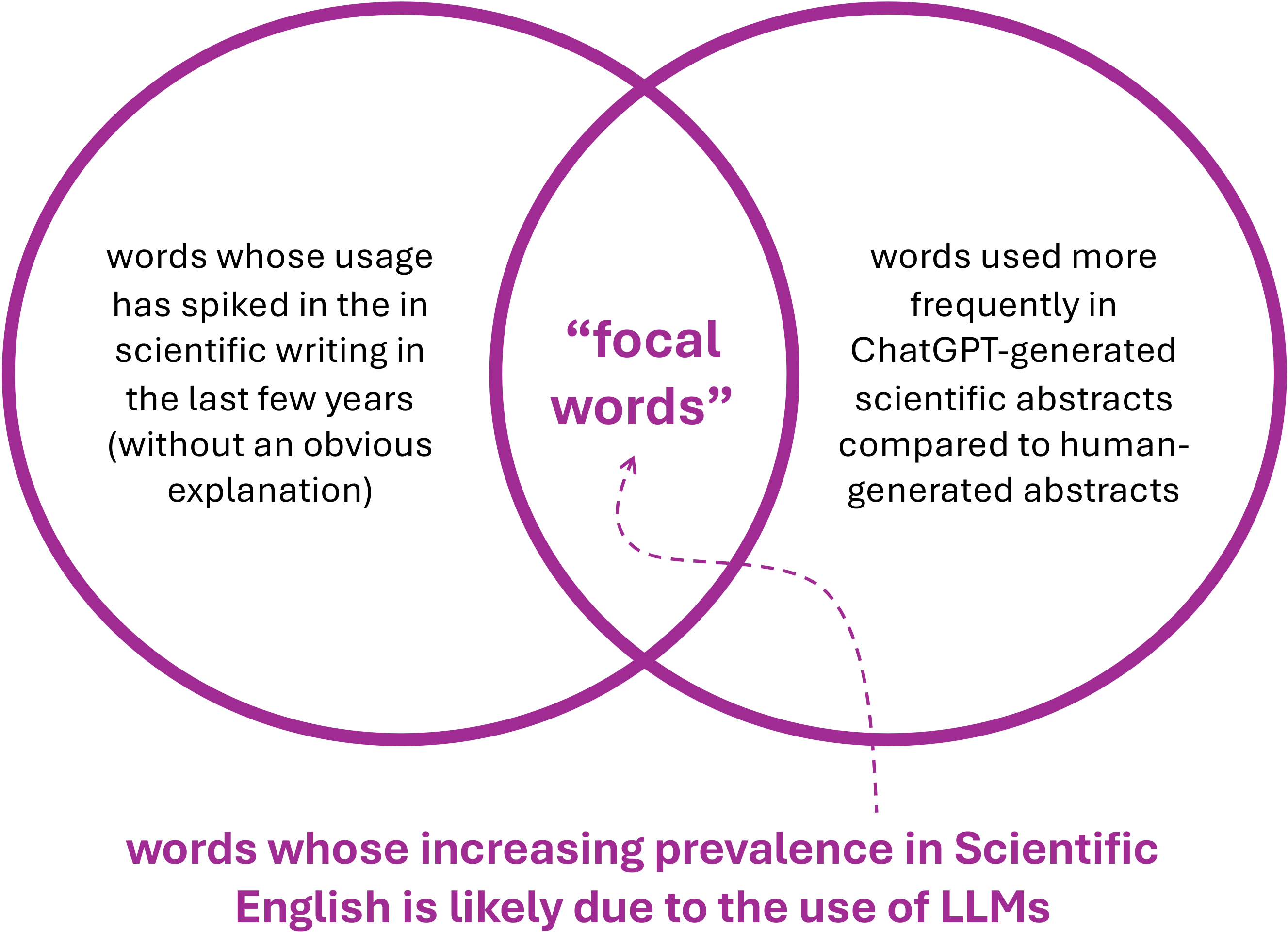}
    \caption{We formalize a procedure for identifying words whose increasing prevalence is likely the result of LLM usage. Although our focus is Scientific English, the method can be applied across domains and languages.}
    \label{fig:venndiagram}
\end{figure}

\noindent Changes in dominant methodological and explanatory frameworks – such as the rise of mechanical philosophy, or the mathematization of scientific fields – have been accompanied by changes in word usage and syntactic structures as well \cite{degaetano2018using,krielke2024cross}. Such changes continue through the present \cite{banks2017extent, leong2020passive}.

Over the last two years, however, Scientific English has witnessed increasing usage of certain lexical items at a seemingly unprecedented pace. Discussions on social media (e.g.,\ \citealt{Koppenburg2024,Nguyen2024,shapira2024delving}) and in academic discourse \cite{gray2024chatgpt,kobak2024delving,liang2024mapping,liu2024towards,matsui2024delving} have pointed out that words such as ``delve,'' ``intricate,'' and ``nuanced'' have appeared far more frequently in scientific abstracts from 2023 and 2024 compared to earlier years. Unlike many previous changes in Scientific English, these trends do not seem to be explained by changes in the content of science or in wider language use. Instead, it is widely assumed that the sharp increase is due to the use of large language models (LLMs) like ChatGPT for scientific writing. Evidence supporting this hunch has recently emerged (e.g.,\ \citealt{cheng2024have,liang2024monitoring}). 

The goals of the present research were twofold. First, we aimed to provide a systematic characterization of this linguistic phenomenon. Some existing work has relied on informal methods to identify words observed to occur more frequently in AI-generated writing (e.g.,\ \citealt{matsui2024delving}). We developed a method for extracting lexical items of interest, described in Section~\ref{sec:corpusanalysis}, which is rigorous, reproducible, and transferable to other data and models. We identified 21 ``focal words'':\ lexical items that have recently spiked in Scientific English and are overused by ChatGPT-3.5 in scientific writing tasks, as illustrated in Figure \ref{fig:venndiagram}.

Prior research has focused on quantifying such focal words' increasing prevalence and estimating how much recent scientific writing has been produced with LLM assistance (e.g.,\ \citealt{kobak2024delving,liang2024mapping}). By contrast, our second goal was to explore the factors that might contribute to the phenomenon of lexical overrepresentation:\ \textit{Why} does ChatGPT use ``delve'' (and other focal words) so frequently when generating scientific text? We identified a set of possible factors, characterized in Section~\ref{sec:puzzle}, and began to assess them. We did not find evidence that model architecture or algorithmic decisions play a major role in the overrepresentation of focal words (Section~\ref{sec:model}), nor that lexical overrepresentation stems from training or fine-tuning data (Section~\ref{sec:training}). 

LLM training often involves reinforcement learning based on information about quality outputs from human evaluators. We found mixed evidence that reinforcement learning from human feedback (RLHF) contributes to the overrepresentation of our focal words in LLM-generated text. Positive evidence comes from model testing on Meta's Llama LLM (Section~\ref{sec:model}). An exploratory experiment described in Section~\ref{sec:rlhf} is inconclusive, although our findings indicate that participants became wary of the word ``delve'' in the first sentence of an abstract (e.g.,\ 'This article delves into ...'). Since the experiment's inconclusiveness stems partly from methodological issues, we believe a follow-up study is warranted. Many important questions about the future of LLM-driven language change remain (Section~\ref{sec:conclusion}).

\section{Corpus Analysis:\ Identification of Overrepresented Lexical Items}
\label{sec:corpusanalysis}

\begin{figure}[ht]
    \centering
    \includegraphics[width=1\columnwidth]{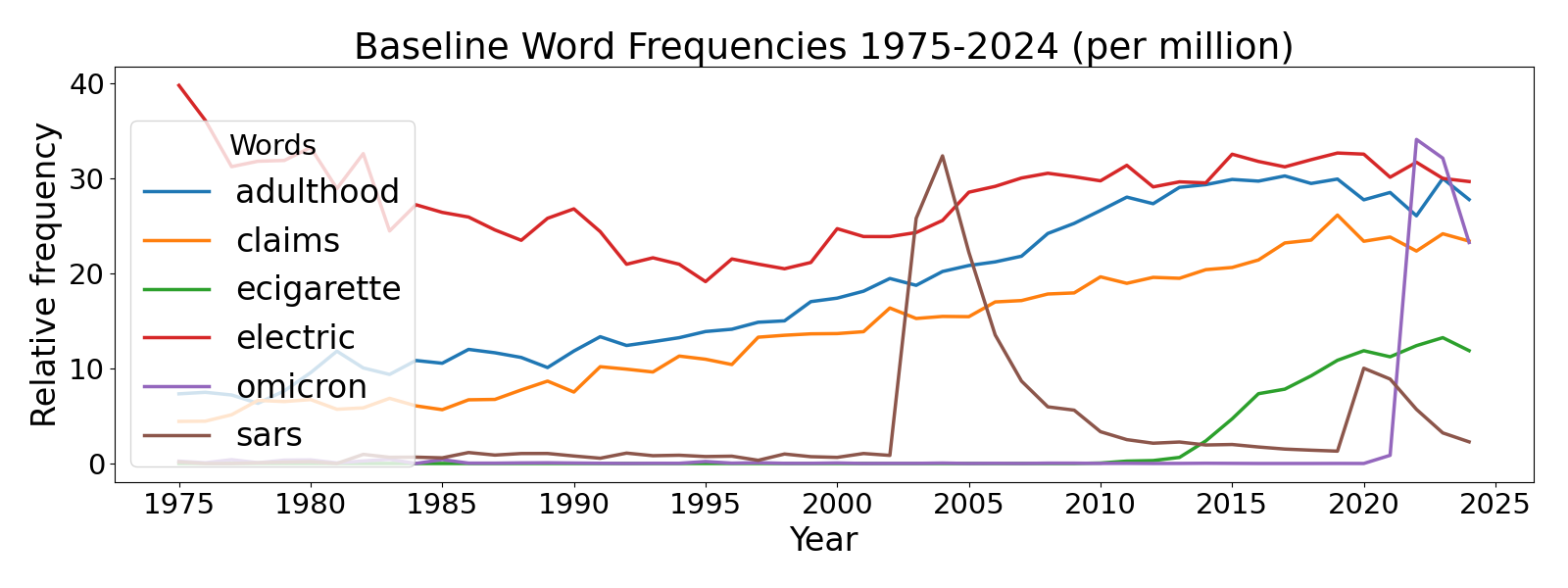}
    \caption{Selected lexical entries:\ change over time.}
    \label{fig:baselineopms}
\end{figure}

To probe recent changes in Scientific English, we used PubMed's publicly available repository of scientific abstracts, which focuses on biomedical literature \cite{PubMed} (downloaded through the PubMed API using a Python script \cite{Python3}; Snapshot:\ May 4, 2024; all code on our GitHub). Our analysis includes more than 5.2~billion tokens (inflected forms) from 26.7 million abstracts. To track changes in word usage over time, we measured occurrences per million (opm) of a given token in each year. Figure~\ref{fig:baselineopms} illustrates the usage trajectories of some baseline items over time. We focus on the period from 1975 to May 2024 as data prior to 1975 are less extensive.

The goal of our corpus analysis was to identify words whose recent overuse in scientific writing is likely the result of LLM deployment. Our approach involved three steps. First, we determined which words were more prevalent in abstracts from 2024 compared to 2020 (since LLMs were not widespread pre-2021). We calculated the percentage increase in opm for each token in the database between 2020 and 2024. Unsurprisingly, there was a straightforward explanation for why some words spiked in usage during that time. For example, ``omicron'' and ``metaverse'' were two of the words that showed the largest percentage increase (for ``omicron'', see Figure~\ref{fig:baselineopms}). We only considered increases deemed significant by chi-square tests, of which there were about 7300. 

We were interested in isolating words whose spike in usage was unexplained. The authors functioned as annotators and independently reviewed the list of words that had the highest percentage change to exclude irrelevant tokens (like year numbers) and words whose spiking had an explanation in terms of scientific advances or world events. In cases of disagreement, we included the word on our list. We stopped once we had 50 words whose usage spiked without any obvious explanation (see incl.ods on GitHub). This list contained several of the words that had been the focus of online conversation, including ``delve'' and ``intricate''.

However, a spike without an obvious explanation is not necessarily LLM-induced. For example, the usage of 'mash' increased tenfold, but it is not a word that ChatGPT is known to overuse. The second step of our method involved identifying words that are overrepresented in AI-generated scientific abstracts compared to human-generated abstracts. In producing AI-generated abstracts, our aim was to imitate the process by which researchers might have deployed an LLM in 2022-early 2024 (while paying attention to careful prompt formulation \cite{wei2022chain,zhou2022least}). After some exploration, we ended up with a two-stage process:\ (1) We randomly sampled 10,000 abstracts from papers published in 2020 from the PubMed database. Via the API, ChatGPT-3.5 then summarized the associated paper (Prompt:\ ``The following is an abstract of an article. Summarize it in a couple of sentences.'') (2) The ChatGPT-generated summary was then used to ask ChatGPT-3.5 for a corresponding scientific abstract. (Prompt:\ ``Please write an abstract for a scientific paper, about 200 words in length, based on the following notes.'') We suspect that the most common way of using an LLM to generate an abstract back when ChatGPT could not accept paper-length inputs involved providing important fragments of a paper. We used ChatGPT-3.5 for the entirety of our project because if scientific abstracts in our dataset contain AI-generated language, it is most likely from ChatGPT-3 or ChatGPT-3.5 \cite{Sarkar2023}.

In total, from 10,000 human abstracts, we generated 9,953 AI abstracts. (For a small number, ChatGPT would not provide a response, presumably due to topic sensitivity.) We then compared the word usage in the AI-generated abstracts with word usage in the original abstracts. We only considered words for which a chi-square test indicated a significant difference in opm between the human- and AI-produced text. This gave us a list of items overused by ChatGPT. 

\begin{figure*}[ht]
    \centering
    \includegraphics[width=1.5\columnwidth]{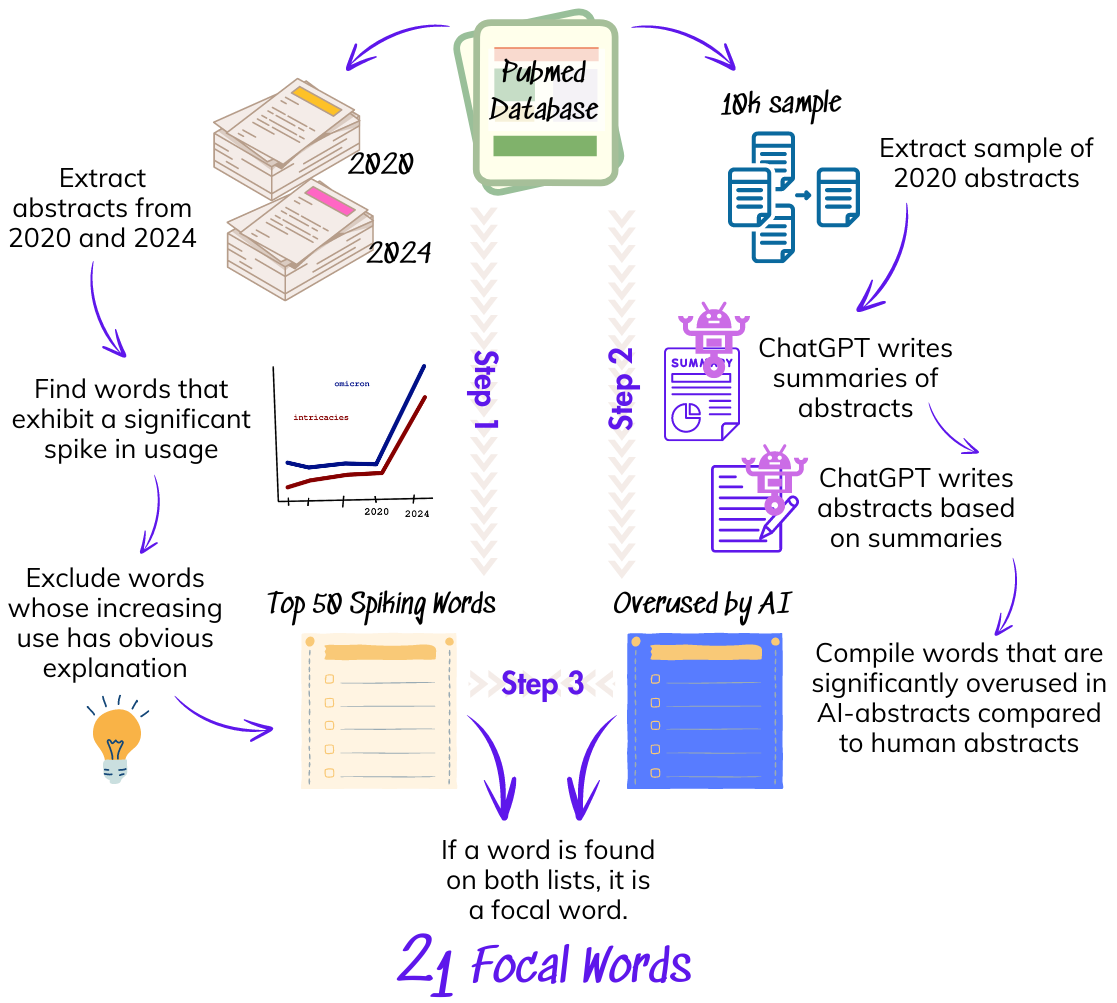}
    \caption{Our method for the systematic identification of focal words.}
    \label{fig:focalwordprocedure}
\end{figure*}

\begin{figure*}[ht]
    \centering
    \includegraphics[width=1.45\columnwidth]{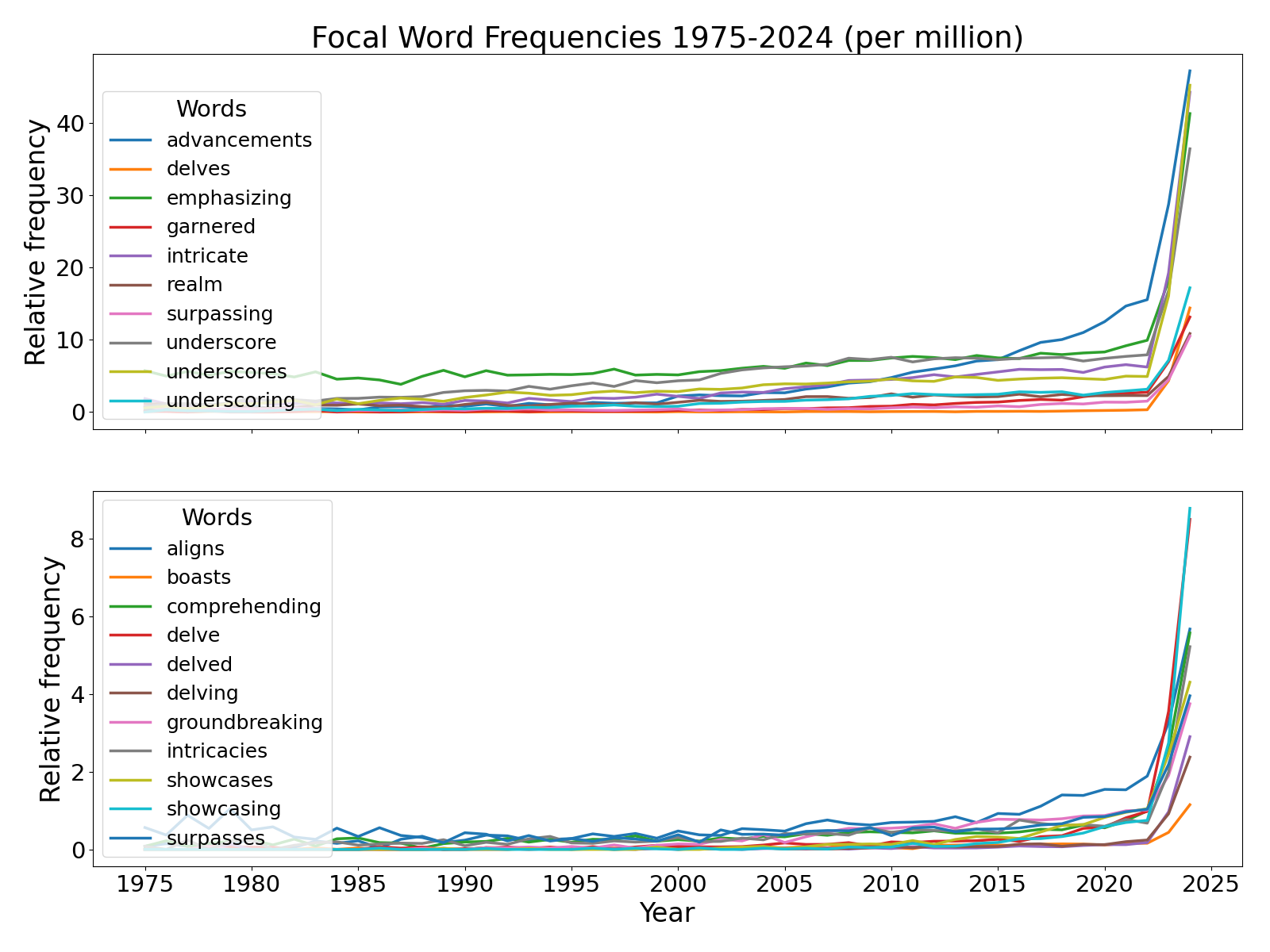}
    \caption{Occurrences per million words in PubMed abstracts for our 21 focal words.}
    \label{fig:focalwordsopms}
\end{figure*}

In the third step of our analysis, we returned to the list of 50 spiking words to ask:\ Is the word also on the ChatGPT-overuse list? If so, then it became a ``focal word'' (Figure~\ref{fig:focalwordprocedure}). This gave us a list of 21 focal words (Figure~\ref{fig:focalwordsopms} and Appendix~\ref{sec:appendix-a}).\footnote{Since the part-of-speech category is not always clear for a given token, the focal word list contains inflected forms instead of lemmata.} Each focal word (a) shows a significant spike in opm between 2020 and 2024, (b) its spike lacks an obvious explanation, and (c) ChatGPT tends to use it significantly more than humans when writing scientific abstracts (Figure \ref{fig:venndiagram}). Thus, a plausible explanation for the increasing prevalence of each focal word in Scientific English is the use of AI.

This systematic, three-step method for identifying focal words is novel. It improves on more informal ways of identifying AI-associated words, and it can be applied to other corpora and LLMs beyond ChatGPT-3.5. (Appendix~\ref{sec:appendix-b} reports similar results for ChatGPT-4.0(-mini).) Future research can use the method to investigate whether the same words are overrepresented in the outputs of different models – or whether there are LLMs that do not exhibit lexical overrepresentation at all.

\section{The Puzzle of Lexical Overrepresentation}
\label{sec:puzzle}

A question now presents itself:\ Why are certain words used so often in AI-generated scientific writing? We call this ``the puzzle of lexical overrepresentation.'' There are a number of factors that might be responsible for the overrepresentation of focal words in scientific abstracts generated by ChatGPT. Importantly, these potential explanations are not mutually exclusive:\ multiple factors may (and probably do) contribute. 

\begin{enumerate}
    \item \textbf{Initial Training Data} Although the focal words are overrepresented relative to human-written abstracts, it is possible that they are not overrepresented relative to the data on which ChatGPT was trained to do next-word prediction. Perhaps these words are actually being used by LLMs with the same frequency as in their training data.

    \item \textbf{Fine-Tuning Training Data} After LLMs have been trained on next-word prediction, they are often fine-tuned. For instance, chatbots are presented with sample dialogues to familiarize them with the structure of a conversation. It is possible that something about ChatGPT's fine-tuning data leads it to favor certain words (e.g.,\ if the focal words are overrepresented in the sample dialogues).

    \item \textbf{Architecture} Another possibility is that there is something about the architecture of LLMs, or perhaps ChatGPT in particular, that causes them to overuse certain words. Maybe LLMs' transformer architecture tends to privilege some lexical items over others in an as-yet-unrecognized way. (Even if this explanation proves correct, the question remains why this particular set of words is overrepresented.)

    \item \textbf{Choice of Algorithms} LLM development involves many different algorithms. Tokenization algorithms, for example, segment an input string into discrete lexical items. It is possible that the choice of one algorithm over others causes lexical overrepresentation. Why the algorithm does so, and why these particular words are overused, would then be further questions.

    \item \textbf{Context Priming} A well-known strength of LLMs is sensitivity to genre. Their outputs are highly dependent on the domain and style requested by the prompt. Perhaps there is something about being asked to write scientifically that causes ChatGPT to overuse the focal words. That is, maybe ChatGPT associates scientific writing in particular with words like ``delve'' and ``intricate.'' This explanation, if correct, raises the further question of why ChatGPT has this association.

    \item \textbf{Reinforcement Learning from Human Feedback (RLHF)} Human feedback is used in later training stages to give LLMs information about what a quality output looks like. A human evaluator might rate several potential responses, for example, with the model then trained with reinforcement learning to produce responses similar to highly-rated exemplars. It is possible that this human feedback encodes a preference for certain words. If responses containing ``delve'' and ``intricate'' are rated more highly by evaluators, it would explain why there is overrepresentation of these words in model outputs.

    \item \textbf{Other factors} This list of potential explanations is not exhaustive. Many other choices – e.g.,\ parameter settings, including temperature, Top K – might influence lexical overrepresentation in LLM outputs.
\end{enumerate}

\noindent Apportioning responsibility for lexical overrepresentation to these factors is not straightforward. The puzzle of lexical overrepresentation arises in part because LLMs are to some extent ``black boxes'' \cite{knight2017dark,sculley2015hidden}. Pending further advances in LLM explainability or interpretability (e.g.,\ \citealt{templeton2024scaling}), we may struggle to understand many aspects of their behavior. An additional obstacle, however, is that many aspects of LLM construction are closely-guarded secrets. Information that would help discriminate between the potential explanations above is not public, even for open source models. For instance, we do not know exactly what data LLMs are trained on (relevant to \#1 above), which fine-tuning steps there are (\#2), what genres the models are exposed to during training (\#5), and who the human evaluators are (\#6). In the remaining sections we pursue several indirect ways of probing potential explanations of the puzzle of lexical overrepresentation.

\section{Searching for Overrepresentation in Possible Training Data}
\label{sec:training}

Our focal words are overrepresented in text generated by ChatGPT compared to earlier PubMed abstracts. Other research indicates that such words also appear less frequently in related datasets in the pre-LLM era \cite{liang2024mapping,liang2024monitoring,gray2024chatgpt}. Although we do not know exactly what data LLMs have been trained on, these results cast doubt on the hypothesis that ChatGPT is using words like ``delve'' and ``surpass'' frequently because they occur frequently in its training data. 

To further demonstrate that the focal words are probably not overrepresented in the training data, we analyzed several additional datasets, namely:\ Arxiv abstracts (accessed 4 Aug 2024; contains data from 1986 onwards, averaged over all years), the Leipzig Corpus Collective (\citealt{goldhahn2012building}; the English LCC contains mostly news texts and transcriptions, data from 2005 onwards; preprocessed snapshot from a previous project), and Wikipedia articles and discussions (\citealt{wikipedia_dump_2024}, accessed 4 Aug 2024). The results are presented in Appendix~\ref{sec:appendix-b}. The opm of the focal words in our ChatGPT-3.5-generated abstracts far exceeds their opm in the four datasets examined.

Second, we conducted a similar analysis for various varieties of English using the International Corpus of English (ICE; \citealt{kirk2018international}). Although ICE is relatively small compared to the other datasets (the subcorpora for most varieties contain about one million words), we do not find evidence that the focal words are especially prevalent in any particular variety of English (see Appendix~\ref{sec:appendix-c}). This suggests that the overrepresentation of focal words in ChatGPT's outputs is probably not due to an overrepresentation of a certain variety of English in its training data. It has been hypothesized that LLMs might frequently use words like ``delve'' because they are more common in varieties of English spoken by human evaluators who provide fine-tuning data, such as Nigerian English \cite{Hern2024}. Our initial analysis of ICE does not support this hypothesis.

\section{Model Choices:\ Architecture and Algorithms}
\label{sec:model}

Could choices about model architecture or algorithms be responsible for the puzzle of lexical overrepresentation? To probe this, we would ideally build an LLM ourselves and test the impact of each potential factor on the prevalence of focal words. This requires vast resources, however, and is beyond most researchers' capabilities, including our own. A more feasible alternative would be to investigate a model that has several released variants – e.g.,\ different versions of the same model using different optimization algorithms. Such a model must also be queryable with respect to information-theoretic measures like entropy \cite{shannon1948mathematical}. To our knowledge, no LLM offers such fine-grained releases. 

 The closest we could find is the comparison between Llama 2-Base (Llama-2-7b-hf) and Llama 2-Chat (Llama-2-7b-chat-hf; \citealt{touvron2023llama}). We used the Llama 2 models because they are more similar to ChatGPT-3.5 than Llama 3 (\citealt{chiang2024chatbot};  but Llama 3 produces similar results; Appendix~\ref{sec:appendix-e}). The main difference between these two versions of Llama is that Llama 2-Chat includes fine-tuning and RLHF, whereas Llama 2-Base does not. Llama models can also be queried for per-word entropy \cite{jurafsky2024slp}. 

\begin{equation}
\label{eq:entropy}
H_{\text{p-w ent}} = -\frac{1}{L} \sum_{i=1}^{n} p(x_i) \log p(x_i)
\end{equation}

By comparing the two models' per-word entropy for human- and AI-generated abstracts, we could assess which was more ``surprised'' by abstracts with an overrepresentation of focal words. Any difference between the models provides evidence about the source of lexical overrepresentation. We provided our sample of 10,000 human-written abstracts to both versions of Llama 2, followed by the abstracts rewritten by ChatGPT-3.5 (see Section~\ref{sec:corpusanalysis}). The results are presented in Table~\ref{tab:llama_comparison}.  				

\begin{table}[ht]
\centering
\begin{tabular}{lcc}
\hline
 & \textbf{Llama 2-Base} & \textbf{Llama 2-Chat} \\
\hline
Human  & 1.616 & 1.051 \\
AI     & 1.633 & 0.886 \\
\hline
\end{tabular}
\caption{Per-word entropy for human abstracts compared to ChatGPT-generated abstracts. Higher values of entropy mean that the model is more ``surprised.''}
\label{tab:llama_comparison}
\end{table}

We observe that Llama 2-Base is slightly less ``surprised'' by human-written text, while Llama 2-Chat is considerably less ``surprised'' by AI-generated abstracts, in which the focal words are overrepresented. This suggests the overuse of focal words might be driven by some factor that differs between the models. Given that model architecture and many algorithms are held constant across Llama 2-Base and Llama 2-Chat, our findings suggest that these factors are not the primary causes of lexical overrepresentation. Instead, they indicate that fine-tuning and RLHF – which differ between the models – might be important contributors. 

These results are necessarily limited. We cannot claim definitively that the observed difference between the models is driven by the prevalence of focal words rather than some other feature of AI-generated text. Moreover, most of our paper is concerned with ChatGPT rather than Llama. The difficulty is that there are no models of ChatGPT (v.3 or above) that can be queried in the described fashion. We think Llama is a useful approximation.

\section{RLHF:\ An Experimental Approach}
\label{sec:rlhf}

Our model testing with Llama suggested that RLHF might contribute to lexical overrepresentation. This hypothesis has intuitive plausibility:\ when human evaluators assess alternative answers to a query, perhaps they are exhibiting a preference for answers containing certain words. Since LLMs are trained to align their answers with human preferences, they would learn to use those words more frequently \cite{christiano2017deep,ziegler2019fine}. To further investigate this potential explanation, we conducted an exploratory online study in which participants indicated whether they preferred scientific abstracts that contained our focal words.

\textbf{Materials}. We randomly sampled shorter PubMed abstracts (70-100 words) from the year 2020 and, with Python and using the OpenAI API, used ChatGPT-3.5 to rewrite them with and without focal words. (Shorter abstracts were used to keep stimuli of a manageable length for participants.) For the focal-word abstracts, the prompt included four randomly selected words from our list of 21 focal words. An example prompt is:\ ``Please write a 100-word abstract for the following scientific paper, using words such as 'delves,' 'underscores,' 'surpasses,' and 'emphasizing':\ [SUMMARY].'' (The summary was generated via the procedure described in Section~\ref{sec:corpusanalysis}.) The script instructed ChatGPT to generate and revise an abstract until it contained at least three focal words. For the no-focal-word abstracts, we used a similar prompt:\ ``Please write a 100-word abstract for the following scientific paper, making sure not to use words such as [list of blockwords]:\ [SUMMARY].'' The blockwords included the 21 focal words plus another 21 words identified using the methodology described in Section \ref{sec:corpusanalysis}. The script prompted ChatGPT to generate and revise an abstract until it contained none of the blockwords. 

We created 200 items, each consisting of one abstract with focal words and one without (for the same paper). We manually filtered out a handful of ungrammatical or nonsensical abstracts. Considerably more than half of the abstracts with focal words included ``delve'' in the first sentence; we call items containing these abstracts ``delve-initial'' items. To compile a bank of 30 critical items, we selected the 15 delve-initial items and the 15 other items with the smallest difference in length between the abstracts with and without focal words. (We capped delve-initial items at 50\% to prevent participants from detecting the study's purpose.) We also constructed 30 pairs of distractor items in the same manner as the critical items, except both abstracts were generated using the no-focal-word prompt. A full list of experimental items can be found on Github, and two examples are in Appendix~\ref{sec:appendix-c}. 

\textbf{Participants.} We used Prolific (\href{https://prolific.com}{prolific.com}) to recruit participants. Public information about the human evaluators employed to provide feedback in RLHF is limited \cite{ouyang2022training,perrigo2023exclusive}, so we recruited 201 participants from India (140 male, 61 female). Average age was 31.3 years (stdev:\ 10.6). We also collected data on self-assessed English proficiency and first languages (see our GitHub). Participants were compensated at an average rate of \$15 per hour. 

\textbf{Task and Exclusions.} The study began with IRB information, followed by task instructions, and then the items. An image of the interface can be found in Appendix~\ref{sec:appendix-f}. Participants evaluated 20 items in total, indicating which abstract they preferred out of the two presented. The first item was a calibration item, followed by (in random order) 5 critical items, 10 distractor items, 2 items checking language abilities, and 2 attention checks (``This is not a real item, please click on the left button'' inserted in the middle of the text). Thus, the proportion of critical items was 25\%. Each time an item was displayed, it was randomly determined which abstract was displayed on the left vs.\ right. If a participant failed one of the attention checks, their data were disregarded. Participants were warned if they were proceeding unrealistically fast (0.25 * (225 ms + 25ms * character length of an item)), and items with excessively fast rating times were excluded from our analysis (following the methodology from \citealt{haussler2017hot}). We also excluded data from participants who completed less than 10 out of the 20 items. After exclusions, we analyzed a total of 1822 ratings, with 1215 ratings for distractor items and 607 ratings for critical items, resulting in each critical item receiving an average of 20.2 ratings (stdev:\ 3.4). Given the study compensation, the high exclusion rate came as a surprise. 

\textbf{Analysis.} Our original plan was to test all 30 critical items together in a chi-square analysis against the distractor items (an approximation of random choices), to assess whether participants preferred abstracts containing focal words. These results are reported below. However, during the generation of the abstracts, we noticed the aforementioned excess of delves in the first sentence and split the critical items into delve-initial items and other items. A lower N per condition and a higher-than-expected exclusion rate left us considerably below the originally estimated sample size from a pre-study power analysis. Thus, we added an exploratory mixed-effects logistic regression model, with rating as the dependent variable and condition as the independent variable, including items as a random effect (rating ~ condition + (1 | item\_id)). Distractor items served as the intercept condition. For delve-initial items and other items, a preference for the focal-word abstract was encoded with 0, and a preference for the no-focal-word abstract with 1. For the distractor items, there are two no-focal-word abstracts, randomly encoded as 0 or 1. 

\begin{figure}[ht]
    \centering
    \includegraphics[width=1\columnwidth]{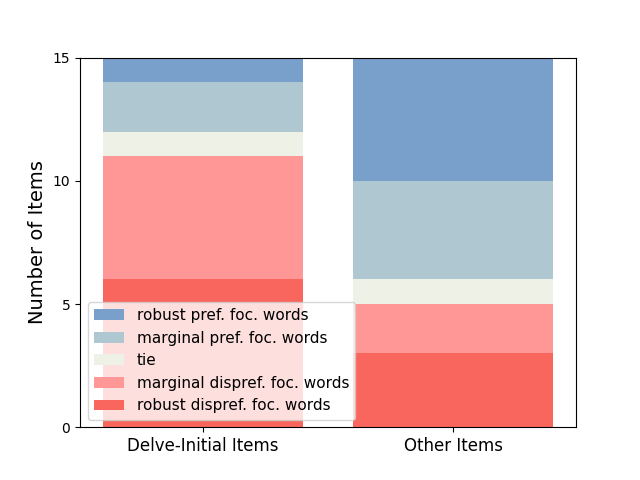}
    \caption{Experimental results:\ Preferences between focal-word and non-focal-word abstracts in delve-initial and other items.}
    \label{fig:experimentalresults}
\end{figure}

\textbf{Results.} Contrary to our expectations, when all critical items are analyzed together, there is a slight preference for the no-focal-word abstracts. However, this overall difference between all critical items and distractor items is not significant in a chi-square test (p = 0.174). The follow-up analysis suggests that this outcome might be driven by the delve-initial items, as Figure~\ref{fig:experimentalresults} illustrates. In the logistic regression model, we observe that the coefficient for the distractor items, represented by the intercept condition, is 0.500 (rounded to the third digit). This indicates that participants did not exhibit a significant preference between the distractor item abstracts, validating our methodology (Appendix~\ref{sec:appendix-g}). The analysis also shows that delve-initial items differed significantly from the distractors (p = 0.023), with a coefficient of 0.082, indicating that for the delve-initial items, participants preferred the abstracts without focal words. Participants exhibited a slight but non-significant preference for abstracts with focal words for the other critical items (coefficient = -0.017; p = 0.651). The group variance was small (0.003), indicating that most of the variability in the ratings was due to the fixed effects. The model converged successfully (log-likelihood = -1324.9522, mean group size = 30.4). A Wald test to determine whether delve-initial items and the other items differed from each other was statistically significant (p = 0.03, Wald Test Statistic:\ 4.77). 

In looking at the responses for each individual item, we consider a preference for the focal-word or no-focal-word abstract of a given pair to be robust if a random outcome falls outside the margin of error, and marginal otherwise (illustrated for the distractors in Appendix~\ref{sec:appendix-g}). This analysis shows a slight difference between delve-initial items and the other critical items:\ participants exhibit a preference for the no-focal-word abstract in more of the delve-initial items, and a preference for the focal-word abstract in more of the other items.

What explains the difference between delve-initial and the other critical items? We suspect that some participants became or were already sensitive to the occurrence of ``delve.'' Participants were probably disproportionately young people with an affinity for technology, and so more likely to be familiar with the discourse surrounding AI language use. Wariness about the word ``delve'' might explain why participants preferred the abstracts without focal words in the delve-initial items (which coincides with a general downturn in sentiment towards LLMs; cf.\ \citealt{leiter2024chatgpt}), though we would like to see these results confirmed with a larger sample.

Having split the critical items in two, a higher N is needed to draw any conclusions about RLHF as a source of lexical overrepresentation, particularly given that we would expect a preference for focal-word abstracts to be subtle. The study warrants a follow-up. We believe that forcing ChatGPT to use certain words when generating abstracts was suboptimal. For example, if an abstract does not initially convey anything about exceeding or outperforming, then a rewritten abstract that includes the focal word 'surpasses' will naturally be worse than the no-focal-word baseline. We suspect that generating critical items in a different way would yield clearer results.

\section{Discussion and Concluding Remarks}
\label{sec:conclusion}

It has been observed that LLMs overuse certain lexical items, a fact even acknowledged by OpenAI \cite{OpenAI2024}. Our work formalized this finding and identified 21 focal words 
whose usage has spiked in scientific abstracts and that are overused by ChatGPT-3.5. These results provide additional evidence that recent changes to Scientific English are partly driven by LLMs. Our work also explored possible explanations of the puzzle of lexical overrepresentation. We failed to find evidence that training data, model architecture, or algorithm choices play a role. However, model testing with Llama was consistent with the hypothesis that RLHF contributes to overuse of particular words by ChatGPT. Our experimental results suggest that human evaluators may treat ``delve'' differently from other focal words.

Future research should further probe the impact of each factor canvassed in Section \ref{sec:puzzle} on lexical overrepresentation. (This includes model choices and training data; despite our negative results, we suspect that these factors do influence the lexical choices of LLMs.) We would especially like to see further work on the role of RLHF. Unfortunately, there are several obstacles to such research, particularly  the lack of procedural and data transparency surrounding LLM development \cite{longpre2024large}. Moreover, it seems that companies building LLMs often solicit feedback from workers who are underpaid, stressed, and under time pressure \cite{toxtli2021quantifying,roberts2022precarious,novick2023dirty}. It is difficult to simulate these conditions ethically in a research environment. Many online recruitment platforms, including Prolific, rightly require decent compensation. 

Although it complicates further study, we think this economic reality lends plausibility to RLHF as a source of lexical overrepresentation. Rushed human evaluators might base their evaluations on the presence of particular words rather than on content, as the former is easier and quicker to evaluate than the latter. If certain words are treated as a proxy for quality, that could explain their overrepresentation in LLM outputs. (We suspect, however, that Scientific English in particular played a minor role in the training of LLMs. It seems more likely that human evaluators rated academic writing in general, with their preferences shaping LLMs' scientific writing through overspill.) This mechanism coheres with our impression that a major social consequence of LLMs is the decoupling of form and content. Many of us take fluency or style as a signal of quality content (\citealt{mcnamara2010linguistic}, and in an L2 context \citealt{kim2018modeling}). Because LLMs are masterful at generating fluid text in just about any style, this heuristic is radically undermined by the increasing ubiquity of LLM-generated text. The irony is that, if our hypothesis about RLHF proves correct, this heuristic has shaped model training as well. LLMs may be undercutting the very same heuristic that has shaped their own lexical preferences. 

It would be interesting to apply our method for identifying focal words to alternative datasets. Although we drew abstracts exclusively from PubMed, future work could examine whether the same focal words have been spiking in scientific disciplines besides biomedicine, in domains beyond Scientific English, and in non-English-language corpora. The method could also be used to probe lexical overrepresentation in LLMs other than ChatGPT. Our impression is that ChatGPT and Llama overuse many of the same words, but a systematic investigation is needed. Finally, additional work on the quirks of LLM-generated language could look beyond the word level \cite{ortmann2021computational}. A virtue of our formalized approach to identifying focal words is that it can be extended in these and any number of other ways to better understand how LLMs are driving linguistic change.

More generally, our research shows that despite the opacity of LLMs, there are ways of probing their behavior and internal workings. Understanding LLMs' linguistic behavior is complicated by their complexity and by secrecy and other industry practices, as mentioned above. Nevertheless, our work indicates that the puzzle of lexical overrepresentation is tractable. Indirect investigative methods can help us explain LLMs' linguistic behavior.

Such research is important because we need to better understand how LLMs are changing language. Almost all of our 21 focal words were already increasing in usage in the years leading up to the release of ChatGPT, suggesting that LLMs may accelerate language change (\citealt{matsui2024delving}; also see \citealt{geng2024impact} and \citealt{yakura2024empirical}). With the increasing prevalence of AI-generated text in many areas of life, LLMs are arguably influencing the language usage even of people who do not themselves interact with these models. Our findings also show that lexical overrepresentation remains a feature of current iterations of ChatGPT (Appendix~\ref{sec:appendix-b}), indicating that the phenomenon is here to stay. 

Still, it is difficult to predict just how AI will shape language in the future. Discussions on social media and in academic discourse, plus our exploratory findings for items with ``delve,'' indicate that there is some public awareness of LLMs' overuse of particular words. This awareness could influence future rounds of RLHF, leading to a realignment of AI and human preferences. At the same time, the language of today – lexical overrepresentations and all – will become the training data for the models of tomorrow, raising concerns about model degradation over time \cite{alemohammad2023self,briesch2023large,hataya2023will,shumailov2023curse}.

One thing is certain:\ through LLMs, tech companies are having a global impact on language usage. We believe this strengthens the case for broader societal debate about the power and responsibilities of these companies. Moreover, our speculations about how the feedback of rushed and underpaid workers might contribute to lexical overrepresentation compound ethical worries about the poor working conditions of tech companies' employees in the Global South \cite{kwet2019digital,gray2024chatgpt,rohde2024broadening}. There are thus both moral and non-moral reasons to apply greater scrutiny to how human feedback is collected and used in the training of LLMs. 

\newpage

\section*{Acknowledgments}

Many thanks to Gordon Erlebacher, Grady Ward, Olmo Zavala Romero, and participants in FSU's SC Artificial Intelligence Seminar for their valuable input on this project. This research was supported by the FSU College of Arts and Sciences Start-up Fund.

\bibliography{custom}

\appendix

\newpage

\section{List Of Focal Words}
\label{sec:appendix-a}

\begin{table}[ht]
\centering
\begin{tabular}{|l|r|r|r|}
\hline
\textbf{Word} & \textbf{opm} & \textbf{opm} & \textbf{Incr.} \\ 
 & \textbf{2020} & \textbf{2024} & \textbf{\%} \\ 
\hline
delves         & 0.21  & 14.38  & 6697.14 \\
delved         & 0.12  & 2.90   & 2240.47 \\
delving        & 0.12  & 2.38   & 1816.83 \\
showcasing     & 0.59  & 8.79   & 1396.03 \\
delve          & 0.58  & 8.50   & 1374.92 \\
boasts         & 0.11  & 1.15   & 918.18  \\
underscores    & 4.50  & 45.19  & 903.61  \\
comprehending  & 0.56  & 5.58   & 898.95  \\
intricacies    & 0.60  & 5.22   & 772.85  \\
surpassing     & 1.37  & 10.50  & 667.48  \\
intricate      & 6.22  & 44.22  & 611.24  \\
underscoring   & 2.70  & 17.17  & 536.94  \\
garnered       & 2.44  & 13.13  & 437.19  \\
showcases      & 0.82  & 4.31   & 422.45  \\
emphasizing    & 8.30  & 41.27  & 397.12  \\
underscore     & 7.42  & 36.40  & 390.65  \\
realm          & 2.25  & 10.85  & 381.10  \\
surpasses      & 0.85  & 3.96   & 367.55  \\
groundbreaking & 0.87  & 3.75   & 330.42  \\
advancements   & 12.49 & 47.17  & 277.59  \\
aligns         & 1.55  & 5.68   & 266.97  \\ \hline
\end{tabular}
\caption{Our 21 focal words.}
\label{tab:word_increase}
\end{table}

\section{Analysis Of Further Corpora and GPT-4o}
\label{sec:appendix-b}

\begin{table*}[ht]
\centering
\begin{tabular}{|l|r|r|r|r|r|r|}
\hline
\textbf{Word} & \textbf{ChatGPT} & \textbf{ChatGPT} & \textbf{Arxiv} & \textbf{LCC} & \textbf{Pubmed} & \textbf{Wiki} \\ 
 & \textbf{3.5} & \textbf{4o-mini} &  &  &  & \\ 
\hline
of            & 45624.84 & 42622.65 & 42842.72 & 27363.47 & 38634.99 & 23116.18 \\
and           & 38889.24 & 32537.79 & 26395.28 & 28488.53 & 39469.96 & 21149.63 \\
the           & 63174.05 & 55111.23 & 72009.63 & 59324.62 & 52139.05 & 53379.32 \\
data          & 978.91   & 1075.59  & 2484.20  & 418.29   & 1734.75  & 142.81   \\
results       & 4074.64  & 3307.32  & 2352.13  & 244.52   & 1722.07  & 95.37    \\
i             & 32.21    & 61.17    & 414.03   & 4715.42  & 214.82   & 8041.61  \\
year          & 78.50    & 61.77    & 37.58    & 1076.29  & 217.25   & 397.61   \\
patients      & 4416.82  & 3936.56  & 48.97    & 131.48   & 4775.73  & 23.04    \\
advancements  & 319.37   & 407.59   & 22.54    & 2.56     & 15.53    & 1.11     \\
aligns        & 6.71     & 19.99    & 6.68     & 1.32     & 1.89     & 0.90     \\
boasts        & 5.37     & 0.61     & 0.43     & 14.11    & 0.16     & 1.48     \\
comprehending & 6.71     & 7.27     & 1.77     & 0.37     & 0.99     & 0.31     \\
delve         & 19.46    & 18.17    & 4.07     & 2.23     & 0.98     & 1.21     \\
delves        & 183.17   & 23.01    & 3.20     & 0.79     & 0.32     & 0.53     \\
delved        & 6.71     & 0.61     & 0.30     & 0.61     & 0.18     & 0.38     \\
delving       & 8.72     & 0.61     & 0.72     & 0.76     & 0.24     & 0.61     \\
emphasizing   & 138.21   & 367.61   & 10.21    & 2.82     & 9.92     & 2.64     \\
garnered      & 20.80    & 173.21   & 4.09     & 4.34     & 2.74     & 4.61     \\
groundbreaking & 38.92   & 17.56    & 2.47     & 5.91     & 1.02     & 2.26     \\
intricate     & 163.04   & 316.14   & 17.87    & 4.79     & 6.22     & 2.13     \\
intricacies   & 15.43    & 27.25    & 1.98     & 1.24     & 0.68     & 0.68     \\
realm         & 10.74    & 54.51    & 11.53    & 9.22     & 2.27     & 8.46     \\
showcases     & 28.85    & 4.24     & 3.19     & 4.65     & 1.05     & 1.46     \\
showcasing    & 30.19    & 58.14    & 5.89     & 5.42     & 0.75     & 1.65     \\
surpasses     & 4.03     & 4.24     & 11.16    & 1.14     & 1.04     & 0.40     \\
surpassing    & 5.37     & 17.56    & 7.61     & 1.66     & 1.51     & 1.42     \\
underscore    & 18.12    & 1365.08  & 5.17     & 1.53     & 7.91     & 0.72     \\
underscores   & 60.39    & 1048.94  & 4.95     & 1.90     & 4.91     & 0.90     \\
underscoring  & 10.06    & 313.71   & 2.57     & 0.66     & 3.15     & 0.20     \\ \hline
\end{tabular}
\caption{Occurrences per million for selected baseline words and our 21 focal words. Results are averaged across all given years of the corpus.}
\label{tab:word_occurrences}
\end{table*}

We used the same summaries from the sample of 10,000 abstracts and used a Python script to generate abstracts using GPT-4o-mini, which were then analyzed as per Section~\ref{sec:corpusanalysis}. The system's role was set as 'You are a world-leading scientist.' and the prompt was the same as that used with GPT-3.5. The results are very similar, with a few exceptions:\ 'boasts' is no longer overused; 'delve' is still overused, but to a lesser extent; and the usage of 'underscore' has increased significantly. These differences could be artifacts of the methodology (the GPT-4-generated abstracts are based on the same GPT-3.5 summaries used in Section~\ref{sec:corpusanalysis}), the consequence of active intervention, RLHF workers responding to overuse, and/or other factors. We also conducted a spot-check with a prompt specifying the role 'You are a helpful assistant.' using GPT-4o. For each role, we generated 500 abstracts and analyzed them. There was no noticeable difference with GPT-4o-mini.

\section{Examples Of Critical Items}
\label{sec:appendix-d}

\subsection{A delve-initial item} 

\textbf{Focal-word abstract:} This study delves into the impacts of maintaining mean arterial blood pressure (MABP) at a specific level during the prehospital and initial hospital phases of treatment for patients with traumatic spinal cord injury (TSCI). The results show a strong correlation between maintaining MABP at 80 mm Hg and improved long-term neurological outcomes. Specifically, prehospital and operating room levels had the greatest effect, while maintaining MABP at 80 mm Hg during the first 2 days in the neurointensive care unit (NICU) was also beneficial. These findings surpass previous knowledge and highlight the advancements in comprehending the relationship between MABP and TSCI recovery. 

\textbf{No-focal-word abstract:} This study explores the impact of maintaining mean arterial blood pressure (MABP) at 80mm Hg during prehospital and initial hospital treatment on long-term neurological outcomes in patients with TSCI. Results showed a significant correlation between higher MABP levels and improved outcomes, with the strongest impact observed in the prehospital and operating room phases. The benefits of maintaining MABP at 80mm Hg were also observed in the first 2 days in the NICU. These findings highlight the crucial role of MABP management in minimizing neurogenic shock-induced damage and emphasize the importance of maintaining adequate blood pressure in TSCI patients.

\subsection{A non delve-initial item} 

\textbf{Focal-word abstract:} This paper showcases a novel approach for targeting and disrupting c-di-GMP signaling pathways in bacteria. By utilizing a c-di-GMP-sequestering peptide (CSP), the researchers have developed a method to bind and inhibit c-di-GMP, a key bacterial second messenger. Through structure-based mutations, a more powerful and compact variant of the CSP has been created, effectively preventing biofilm formation in Pseudomonas aeruginosa. This advancement holds promise for controlling bacterial behaviors mediated by c-di-GMP and could have implications for the development of new antibacterial strategies. The results of this study highlight the potential of CSP as a tool for delving into the intricate mechanisms of c-di-GMP signaling.

\textbf{No-focal-word abstract:} A novel approach has been devised for blocking c-di-GMP signaling pathways, a crucial mechanism in bacterial cell functioning. The technique employs a c-di-GMP-sequestering peptide (CSP) that exhibits strong affinity for c-di-GMP and effectively inhibits its signaling. Through targeted mutations, a potent, shortened variant of CSP has been developed, demonstrating efficient inhibition of biofilm formation in Pseudomonas aeruginosa. This innovative method provides a highly promising strategy for targeting c-di-GMP and holds potential for combating various bacterial infections. Further studies could focus on developing more potent and specific CSP variants to fully comprehend and utilize the role of c-di-GMP in regulating bacterial functions.

\section{Per-word Entropy for various Llama Models}
\label{sec:appendix-e}

We validate our results for various Llama models. We use the latest versions available on 28 August 2024. We intend to extend our analysis to the larger, 70bn parameter models. However, due to quota restrictions, we are unable to perform these calculations at this time. We expect that the results will be similar and plan to include them once the account limitations are resolved. 

All models show a drop in average per-word entropy for human input when comparing base models to chat models, with a more pronounced drop observed for AI input. Most models show lower entropy values for human text with the base model compared to AI text. This pattern reverses in the chat models, where AI text shows lower entropy values than human text.

\begin{table}[ht]
\centering
\begin{tabular}{lcc}
%

\hline
8b & \textbf{Llama 3-Base} & \textbf{Llama 3-Chat} \\
\hline
Human  & 1.862 & 1.174 \\
AI     & 1.928 & 1.165 \\
\hline

 & & \\

%

\hline
8b & \textbf{Llama 3.1-Base} & \textbf{Llama 3.1-Chat} \\
\hline
Human  & 1.854 & 1.731 \\
AI     & 1.838 & 1.653 \\
\hline

%

\end{tabular}
\label{tab:llama2_70_comparison}
\end{table}

\FloatBarrier

\section{The Rating Interface}
\label{sec:appendix-f}

\FloatBarrier

\begin{figure}[H]
    \centering
    \includegraphics[width=1\columnwidth]{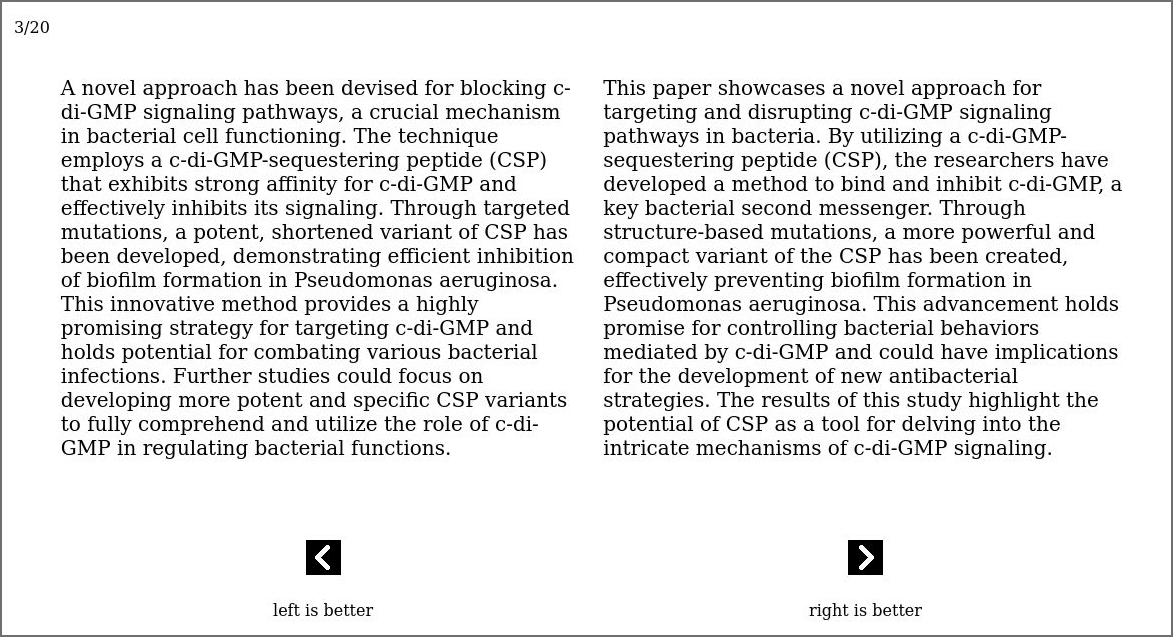}
    \caption{The rating interface for our experiment.}
    \label{fig:interface}
\end{figure}

\section{Ratings For The Distractor Items}
\label{sec:appendix-g}

\begin{figure}[ht]
    \centering
    \includegraphics[width=1\columnwidth]{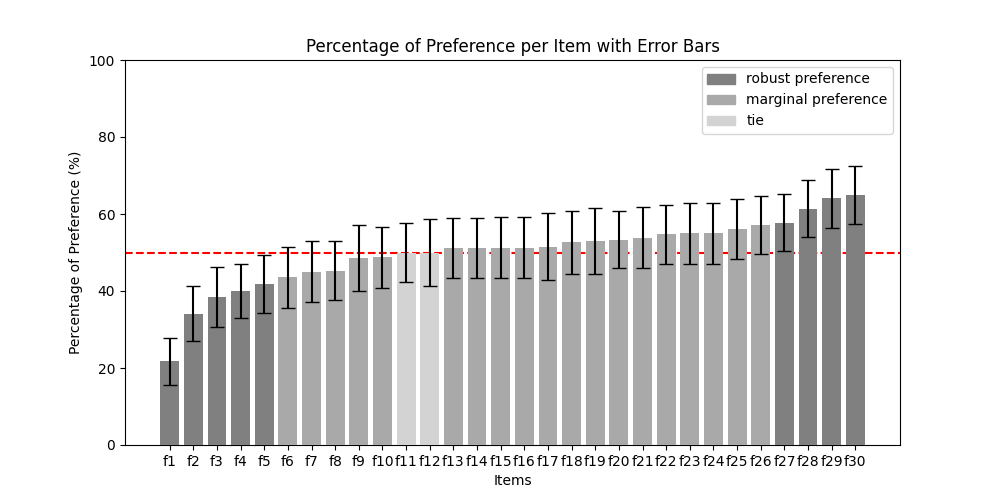}
    \caption{The experimental results for individual distractor items.}
    \label{fig:distractoritems}
\end{figure}

\section{Analysis of the International Corpus of English}
\label{sec:appendix-c}

\begin{figure*}[ht]
    \centering
    \includegraphics[width=2.2\columnwidth]{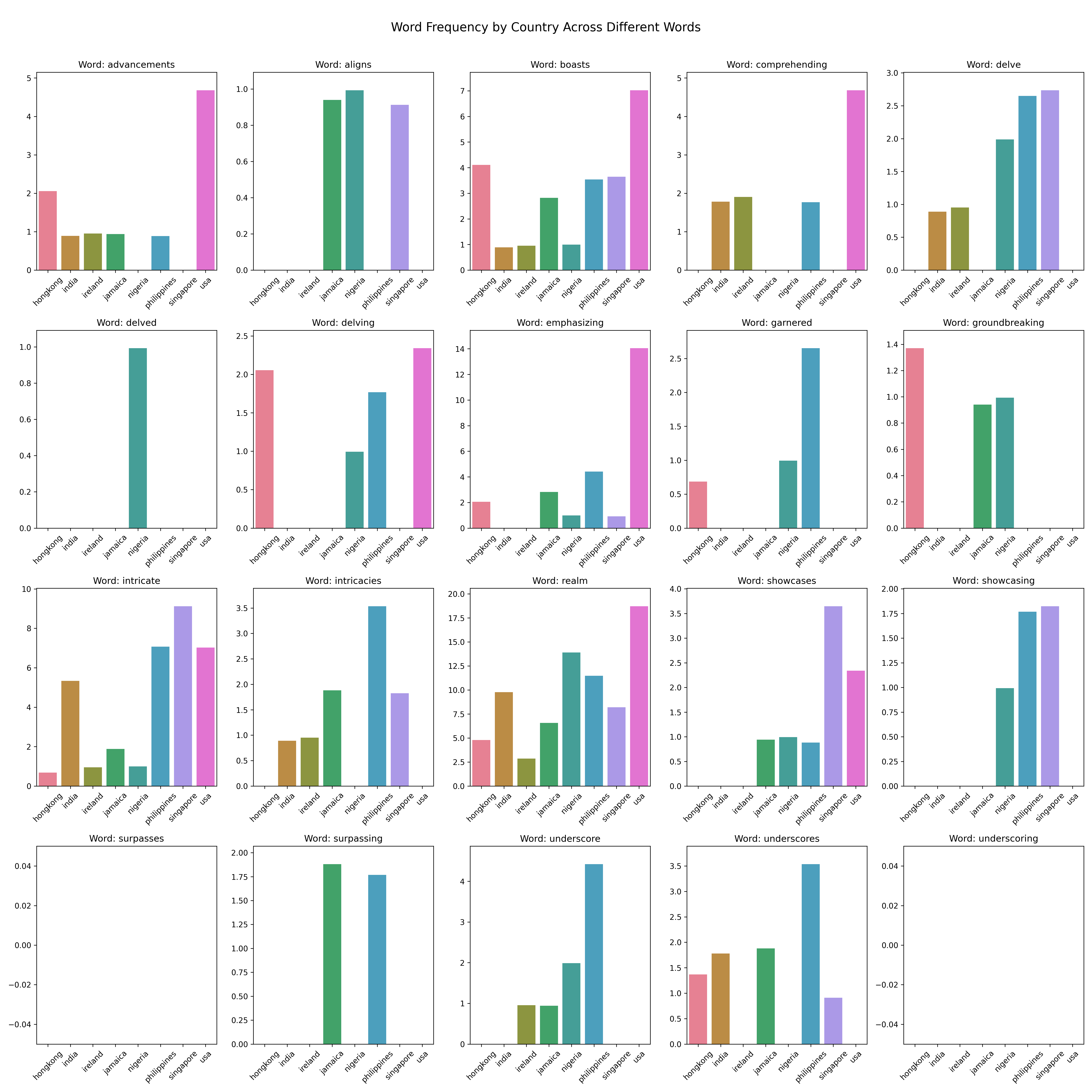}
    \caption{Word frequencies for selected lexical items across various English variants.}
    \label{fig:ice}
\end{figure*}

An analysis of the Englishes of the world can be found in Figure~\ref{fig:ice}. 

\end{document}